\documentclass{article}
\usepackage{ijcai25}

\usepackage{times}
\usepackage{soul}
\usepackage{url}
\usepackage[hidelinks]{hyperref}
\usepackage[utf8]{inputenc}
\usepackage[small]{caption}
\usepackage{graphicx}
\usepackage{amsmath}
\usepackage{amsthm}
\usepackage{booktabs}
\usepackage{algorithm}
\usepackage{algorithmic}
\usepackage[switch]{lineno}

\usepackage{amsfonts}
\usepackage{multirow}
\usepackage[switch]{lineno}
\usepackage{subfig}
\usepackage[labelformat=simple]{caption}
\usepackage{pdfpages}
\usepackage{booktabs} 

\title{Parameter-Efficient Fine-Tuning with Circulant and Diagonal Vectors}

\author{
Xinyu Ding \and
Lexuan Chen \and
Siyu Liao \And
Zhongfeng Wang \\
\affiliations
Sun Yat-sen University\\
\emails
dingbai1357718507@gmail.com,
chenlx77@mail2.sysu.edu.cn,
liaocs2008@gmail.com,
wangzf83@mail.sysu.edu.cn
}

\begin{document}

\maketitle

\begin{abstract}
Foundation models have achieved tremendous success in different domains.
However, their huge computation and storage complexity make these models difficult to fine-tune and also less applicable in practice. 
Recent study shows training in Fourier domain can be an effective fine-tuning method in terms of both model performance and number of training parameters. 
In this work, we propose to further reduce the complexity by the factorization through the product of interleaved circulant and diagonal matrices. In addition, we address the case of non-square fine-tuning weights by partitioning the circulant matrix into blocks.
Our method avoids the construction of weight change matrix and utilizes 1D fast Fourier transform (FFT) instead of 2D FFT. 
Experimental results show that our method achieves similar or better performance across various tasks with much less floating-point operations (FLOPs) and the number of trainable parameters. 
\end{abstract}

\section{Introduction}
Large foundation models (LFMs) are widely utilized in various fields, including natural language processing \cite{KA}, image recognition and generation \cite{CosmicMan}, medical diagnosis \cite{Healthcare}, and autonomous driving \cite{driving}. 
\cite{BERT} have proposed the bidirectional transformer architecture that understands input data from left to right and right to left.
It is trained to predict missing words given input context, and it has served as a foundation model that can be fine-tuned for many downstream tasks. 
Following the transformer architecture, generative pre-trained transformer (GPT) model by \cite{Radford2018ImprovingLU} handles input data from left to right following a sequential prediction order. 
This mechanism turns out successful in many generation tasks such as text summary, question answering, etc.

Although LFMs learn extensive general knowledge during the pre-training phase, they still require extra adjustments in downstream applications to effectively fullfill the task.
Fine-tuning is a typical approach to continue learning on given downstream data and update from pre-trained model parameters. 
While fine-tuning significantly reduces computational costs compared to training from scratch, existing fine-tuning methods still suffer from the huge complexity of LFMs.
As the original model parameters are still kept and maintained during fine-tuning stage, this leaves limited space for the development of fine-tuning methods. 

To address the challenge of fine-tuning LFMs, \cite{LoRA} have proposed low-rank adaptation (LoRA).
This method is an efficient fine-tuning approach designed for LFMs, reducing the number of parameters required during fine-tuning by introducing low-rank matrices. 
The essential idea is assuming the weight change matrix with low rank structure, expressing it as the product of two low rank matrices and only training these two smaller matrices while keeping the original weights frozen.
FourierFT proposed by \cite{FourierFT} assumes a sparse structure in fourier domain of the weight matrix updates \(\boldsymbol{\Delta} \mathbf{W}\).
Although FourierFT reduces the number of training parameters, the computational and storage requirements of the model remain very high, particularly when dealing with LFMs. 
The two-dimensional Fourier transform used to restore \(\boldsymbol{\Delta} \mathbf{W}\) contributes to most of its  computation and storage complexity. 
As a result, the fine-tuning model continues to necessitate high-performance hardware support, including substantial GPU resources and memory, which may be challenging to achieve in practical applications.

\cite{Huhtanen2015FactoringMI} has demonstrated that a general complex matrix \(\mathbf{X} \in \mathbb{C}^{n \times n} \) can be factorized into the product of multiple circulant matrices and diagonal matrices, with total number of factors not exceeding \( 2n-1 \). 
This decomposition method offers several advantages, particularly in terms of computational efficiency and storage optimization. 
The computation and storage of diagonal matrices can be efficiently managed using vector representations. 
Moreover, circulant matrices possess a unique structure that allows them to be diagonalized using the fast Fourier transform (FFT), significantly reducing the complexity of matrix operations and accelerating computation speed. 

Inspired by previous works, we propose circulant and diagonal vector based fine-tuning (CDVFT), which is also a Fourier domain based method. 
Our method represents the weight change matrix \(\boldsymbol{\Delta} \mathbf{W}\) with the product of interleaved circulant and diagonal matrices. 
This factorization simplifies the matrix calculation process and reduces storage requirements. 
Due to the unique properties of matrix product for circulant and diagonal matrices, the quadratic computation complexity now becomes loglinear. 
Different from FourierFT based on 2D FFT, our fine-tuning process avoids the restoration of the weight change \(\boldsymbol{\Delta} \mathbf{W}\) and only takes 1D FFT operations. 
However, the product of interleaved circulant and diagonal matrices inherently forms only square matrices, limiting its generality. To address this issue, we propose partitioning the circulant matrix into blocks, enabling more efficient storage management while maintaining the accuracy standards required for large-scale models.
As a result, CDVFT can achieve efficient storage and computation  at the same time. 
We summarize our main contributions as following:

\begin{itemize} 
\item We introduce CDVFT method that represents \(\boldsymbol{\Delta} \mathbf{W}\) using the product of interleaved block circulant and diagonal matrices. These matrices have linear storage complexity as each of them can be determined by a single weight vector. In practice, we find only a few matrices and a small number of blocks is sufficient to perform fine-tuning.
\item CDVFT avoids the restoration of weight change matrix and has loglinear computation complexity. The circulant matrix vector product can be transformed into 1D FFT, and diagonal matrix vector product is linear in nature. Thus, the overall computation complexity becomes loglinear. 
\item We evaluate our method on natural language understanding, image classification, and instruction and task adjustment. Experimental results show that our method achieves similar or even better results in terms of moder performance, number of training parameters and FLOPs. For example, for the RoBERTa base model, our method results in 51.81$\times$ FLOPs reduction compared to FourierFT and 5.33$\times$ trainable parameters saving compared to LoRA, while resulting in similar or even better accuracy.
\end{itemize}

\section{Related Works}
Fine-tuning LFMs is a challenging problem due to the large model size and computation requirement. 
Although training LFMs from scratch is performed on cloud platforms like LLaMA model by \cite{llama}, fine-tuning is often limited to a specific task and a low-cost computing environment. 
Besides, fine-tuning runs on a much smaller dataset than the pre-training dataset for LFMs. 
Thus, fine-tuning process is expected to be cost-effective. 
The overall complexity should be small and affordable in practice. 

Full fine-tuning is a classical approach training and updating all model parameters at the same time. 
However, it is difficult to perform full fine-tuning on LFMs given the huge computation and storage requirement. 
\cite{NEURIPS2020_1457c0d6} find LFMs are able to generalize to new tasks with few-shot demonstrations as prompt, thereby saving the effort of training on parameters. 
\cite{Prefix-Tuning} argue that adding few-shot demonstrations is bounded by the input length constraint of current LFMs. 
Instead, they propose the prefix tuning method to train a parameter vector and prepend to input,  which is expected to work as prompt in unlimited length.

Updating all model parameters is not desirable in practice, since each task needs to maintain a model. 
\cite{houlsby2019parameter} propose the adapter method, where task dependent parameters are inserted to LFMs.
Fine-tuning process only updates those new parameters, thereby each task effectively sharing pre-trained LFM parameters. 
\cite{mahabadi2021parameter} further reduce the task dependent parameters amount by grouping adapters into a hyper network model such that the network can produce task specific parameters on the fly. 
\cite{sung2022lst} discover backpropagation process through LFMs takes a lot of memory and propose a ladder style adapter design that significantly saves memory consumption. 
Given that adapters bring in extra inference latency due to their new parameters, \cite{lei2023conditional} believe different tasks have different needs for the shared LFM architecture. 
They decide to learn to skip computations in LFM for different adapters, resulting in a faster inference speed.

It can be noticed that adapter adds task dependent parameters and incurs inference delay. 
There are also studies working on mergeable adapters so that after fine-tuning they can be merged into LFM architecture without adding inference latency. 
The essential idea is setting adapter parameters in the same shape as LFM pre-trained parameters, and fine-tuning learns the change of weight parameters, i.e., \(\boldsymbol{\Delta} \mathbf{W}\).  
\cite{LoRA} develop LoRA technique that enforces low rank structure into the weight change matrix. 
Given that LoRA rank can be different for different tasks, \cite{zhang2023adaptive} decide to learn the rank setting by modifying singular values based on importance score function. 
Instead of directly learning on  \(\boldsymbol{\Delta} \mathbf{W}\), 
\cite{FourierFT} propose FourierFT to learn sparse parameters in fourier domain and reconstruct the weight difference using 2D FFT operation. 
It turns out this method requires much less number of parameters, but its reconstruction needs more memory.  

Following the parameter efficient fine-tuning (PEFT) discovery in fourier domain, it is important to look for a method involving matrix and efficient FFT operation. 
Circulant matrix is related to 1D FFT since circulant matrix vector product can be executed using 1D FFT to accelerate. 
There are some studies applying circulant matrix to compress neural networks, such as circulant convolution neural network by \cite{cheng2015exploration} and circulant long short-memory by \cite{wang2018c}. 
However, these works are lack of flexibility on increasing parameter amount and theoretical guarantee on dense matrix approximation. 
\cite{Huhtanen2015FactoringMI} has demonstrated that a general complex matrix \( \mathbf{X} \in \mathbb{C}^{n \times n} \) can be expressed as the product of interleaved circulant and diagonal matrices, with the number of factors not exceeding \(2n-1\):
\begin{equation}
\begin{split}
\mathbf{X} =& \mathbf{A}_{2n-1} \times \mathbf{C}_{2n-2} \times \ldots \times \mathbf{C}_{2j} \times\mathbf{A}_{2j-1} \times \\ 
&\times \ldots \times \mathbf{A}_3 \times \mathbf{C}_2 \times \mathbf{A}_1 \times \mathbf{x}
\end{split}
\end{equation}
where for $j \in \{1,\dots, n\}$, \(\mathbf{A}_{2j-1}\) and \(\mathbf{C}_{2j}\) are diagonal and circulant matrices, respectively. 
Thus, this decomposition theoretically can approximate any dense matrix, and it also enables control on parameter amount by setting number of factors.

\section{Method}

\begin{figure*}[htp]
\centering
\includegraphics[width=\textwidth]{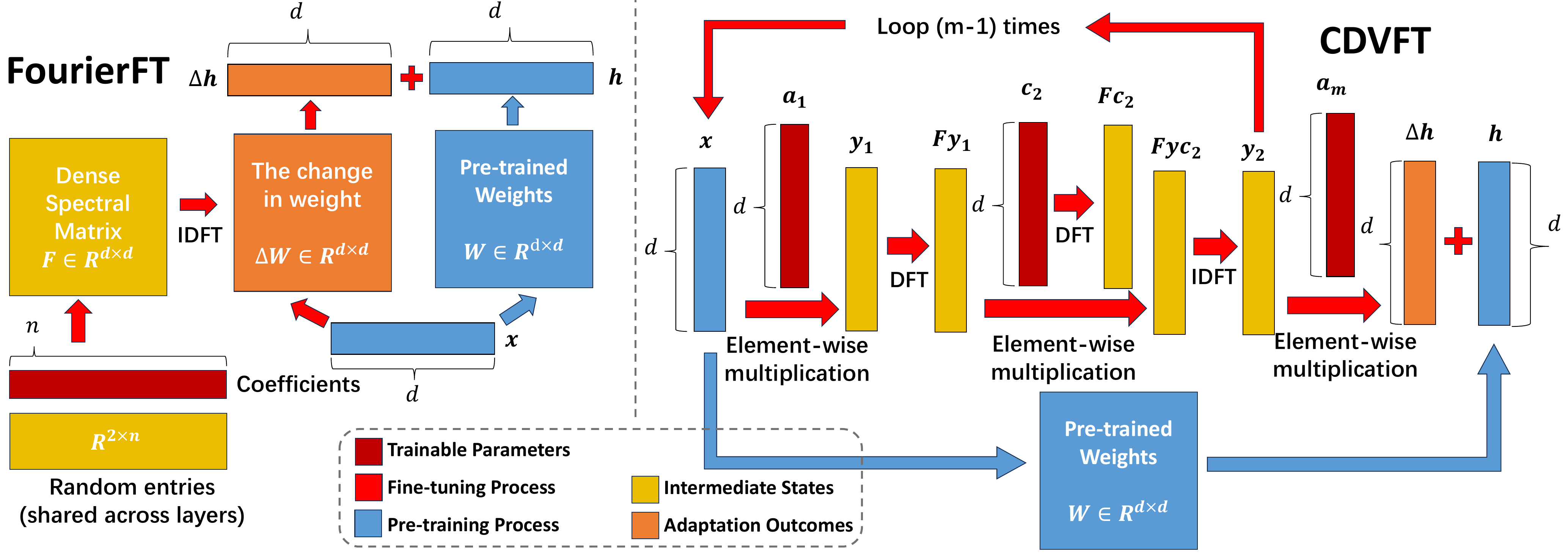}
\caption{
\textbf{Overview of FourierFT (left) and our CDVFT (right).} In FourierFT, one coefficient vector \( \mathbf{c} \in \mathbb{R}^{n} \)  is trained, and it is used to construct the weight change \(\boldsymbol{\Delta} \mathbf{W}\) through 2D FFT operation. In contrast, our CDVFT avoids the construction of \(\boldsymbol{\Delta} \mathbf{W}\), where matrix vector products are transformed into vector operations, i.e., element-wise product and 1D FFT,  significantly reducing computation complexity and memory requirement. In practice, we find $m=1$ (no loops required) can effectively fine-tune the model, where there are two diagonal matrices and one circulant matrix.}
\label{fig:overview}
\end{figure*}

In this section, we introduce circulant and diagonal vector based fine-tuning (CDVFT) method, which is a mergeable adapter design similar to FourierFT. 
After fine-tuning, our trained circulant and diagonal vectors can be used to build circulant and diagonal matrices, which are further combined to reconstruct \(\boldsymbol{\Delta} \mathbf{W}\) and merged into LFMs. 
However, most importantly, CDVFT does not need to recover \(\boldsymbol{\Delta} \mathbf{W}\) during real fine-tuning process, since the reconstruction results in high computation and storage complexity.
Instead, our method takes advantage of the fast matrix multiplication algorithm from circulant and diagonal matrices involving 1D FFT and element-wise product to achieve the goal.

The overall computation flow is illustrated in Fig.\ref{fig:overview}.
It can be seen that CDVFT only takes vector operations at each step, thereby significantly reducing the computation and storage complexity. 
Specifically, according to the findings by \cite{Huhtanen2015FactoringMI} and the unique properties of circulant and diagonal matrix operations, CDVFT first initializes corresponding vectors to represent these matrices. 
It then directly performs multiple element-wise multiplication and 1D FFT on input \(\mathbf{x}\). 
Finally, it yields the output \(\boldsymbol{\Delta} \mathbf{h}\), which can be added to the output  \(\mathbf{h}\) from the original weight matrix  \(\mathbf{W}\).

\subsection{Forward Step}

Let \(\mathbf{x} \in \mathbb{R}^{d \times 1}\) be an input column vector.
Assume weight change matrix \(\boldsymbol{\Delta} \mathbf{W}\in \mathbb{R}^{d \times d}\) that can be decomposed into $2m-1$ factors with $m\leq d$.
Thus, there are $m$ diagonal matrices and $m-1$ circulant matrices. 
For \(j \in \{1, 2, \ldots, m\}\), each diagonal matrix is defined by a vector $\mathbf{a}_{2j-1}\in \mathbb{R}^{d \times 1}$, and each circulant matrix is defined by a vector $\mathbf{c}_{2j}\in \mathbb{R}^{d \times 1}$.
More specifically, they can be expressed as following:
\begin{equation}
\label{eq:diag_circ}
\begin{split}
diag(\mathbf{a}_{2j-1})
&=
\begin{bmatrix}
    \mathbf{a}_{2j-1}^1 & 0  & \dots  & 0 \\
    0 & \ddots  & \ddots & \vdots \\
    \vdots & \ddots & \ddots & 0 \\
    0 & \dots & 0  & \mathbf{a}_{2j-1}^d
\end{bmatrix}
, \\ 
circ(\mathbf{a}_{2j})
&=
\begin{bmatrix}
    \mathbf{a}_{2j}^1 & \mathbf{a}_{2j}^d  & \dots  & \mathbf{a}_{2j}^2 \\
    \mathbf{a}_{2j}^2 & \ddots  & \ddots & \vdots \\
    \vdots & \ddots & \ddots & \mathbf{a}_{2j}^d \\
    \mathbf{a}_{2j}^d & \dots & \mathbf{a}_{2j}^2  & \mathbf{a}_{2j}^1
\end{bmatrix},
\end{split}
\end{equation}
where $diag(\cdot)$ and $circ(\cdot)$ construct a diagonal matrix and circulant matrix, respectively. 
Therefore, the weight change matrix can be written as:
\begin{align}
\begin{split}
\boldsymbol{\Delta} \mathbf{W}
=&
\mathbf{A}_{2m-1} \times \mathbf{C}_{2m-2} \times \mathbf{A}_{2m-3} \times \dots \times  \mathbf{A}_1 
\\
=&
diag(\mathbf{a}_{2m-1}) \times circ(\mathbf{a}_{2m-2}) \times diag(\mathbf{a}_{2m-3}) \\
&\times \dots \times diag(\mathbf{a}_1)
,
\end{split}
\end{align}
where $\times$ is the inner product operation. 
The end-to-end computation flow then becomes:
\begin{equation}
\mathbf{h}' =
\mathbf{h} + \boldsymbol{\Delta} \mathbf{h}
=
\mathbf{W} \times \mathbf{x} + \alpha \times \boldsymbol{\Delta} \mathbf{W} \times \mathbf{x}
\label{eq.5}
,
\end{equation}
where $\alpha$ is a hyper-parameter scalar as in LoRA \cite{LoRA},  $\mathbf{W}\in\mathbb{R}^{d\times d}$ is the pre-trained weight matrix in given LFM and $\mathbf{h}'$ is the new output after  adding our CDVFT adapters. 
This can also be seen in Fig. \ref{fig:overview}. 

We perform the computation from rightmost to leftmost, thereby avoiding the reconstruction of $\boldsymbol{\Delta} \mathbf{W}$ during fine-tuning process.
Let \(\mathbf{y}\in \mathbb{R}^{d \times 1}\) represent the intermediate calculation result from matrix vector multiplications.
Thus,  \(\mathbf{y}_{2j-1}\) is the result from diagonal matrix vector  multiplication, and \(\mathbf{y}_{2j}\) is the result from circulant matrix vector multiplication. 
Note that diagonal matrix vector product is equivalent to element wise product of \(\mathbf{a}_{2j-1}\) and input vector:
\begin{equation}
\begin{split}
&\boldsymbol{\Delta} \mathbf{W} \times \mathbf{x} = \mathbf{A}_{2m-1}  \times \ldots \times \\
&\ \ \ \ \times \overbrace{\mathbf{C}_{2j} \times \underbrace{\mathbf{A}_{2j-1} \times \ldots \times \mathbf{A}_3 \times \mathbf{C}_2 \times \mathbf{A}_1 \times \mathbf{x}}_{\mathbf{y}_{2j-1}}}^{\mathbf{y}_{2j}} \label{eq.4}
,
\end{split}
\end{equation}
\begin{equation}
\begin{split}
\mathbf{y}_0 = \mathbf{x}
,
\ \ 
\mathbf{y}_{2j} = \mathbf{C}_{2j} \times \mathbf{y}_{2j-1},
\end{split}
\end{equation}
\begin{equation}
\begin{split}
\mathbf{y}_{2j-1} = \mathbf{A}_{2j-1} \times \mathbf{y}_{2j-2} = \mathbf{a}_{2j-1} \odot \mathbf{y}_{2j-2}  \label{eq.2} 
,
\end{split}
\end{equation}
where $\odot$ means the element-wise product. 
The circulant matrix vector product can be transformed into 1D FFT operations:
\begin{equation}
\label{eq:circ_matvec}
\begin{split}
\mathbf{F}_{2j-1} &= \text{FFT}(\mathbf{y}_{2j-1}) = \{\sum_{q=0}^{d-1} \mathbf{y}_{2j-1}^{q} e^{-i2\pi \frac{p}{d} q}\}_{p=0}^{d-1}
, 
\\
\mathbf{F}_{2j}&= \text{FFT}(\mathbf{c}_{2j}) =\{\sum_{q=0}^{d-1} \mathbf{c}_{2j}^{q} e^{-i2\pi \frac{p}{d} q}\}_{p=0}^{d-1}
,
\\
\hat{\mathbf{F}} &= \mathbf{F}_{2j-1} \odot \mathbf{F}_{2j}
, 
\\
\mathbf{y}_{2j} 
&= \text{IFFT}(\mathbf{F})
= \{\frac{1}{d} \sum_{q=0}^{d-1} \hat{\mathbf{F}}^{q}e^{i2\pi \frac{p}{d} q}\}_{p=0}^{d-1}
,
\end{split}
\end{equation}
 where \(e^{i2\pi \frac{p}{d} q}\) is the constant term in the Fourier transform, \(i\) indicates the imaginary unit, and \(p\) is the frequency index of the transform.
 We use letter $\mathbf{F}$ to indicate vectors in fourier domain. 
\(\mathbf{F}_{2j-1}\) and \(\mathbf{F}_{2j}\) represent the Fourier transform results of \(\mathbf{y}_{2j-1}\) and the circulant matrix vector \(\mathbf{c}_{2j}\), respectively. 
 \(\hat{\mathbf{F}}\) is the result of element wise multiplication of \(\mathbf{F}_{2j-1}\) and \(\mathbf{F}_{2j}\). 
In consequence, \(\mathbf{y}_{2j}\) is the result of inverse fast Fourier transform (IFFT) of \(\hat{\mathbf{F}}\).

\subsection{Backward Step}
Following current deep learning design, we provide the gradient calculation with respect to $\mathbf{a}_{2j-1}$ and $\mathbf{c}_{2j}$ for all $j$. 
Denote the objective function (i.e., loss function) as \(\mathcal{L}(\cdot)\). 
The backpropagation follows the chain rule, and we can get:
\begin{align}
\frac{\partial \mathcal{L}}{\partial \mathbf{a}_{2j-1}}
=
\frac{\partial \mathcal{L}}{\partial \mathbf{y}_{2j-1}}\frac{\partial \mathbf{y}_{2j-1}}{\partial \mathbf{a}_{2j-1}}
=
\frac{\partial \mathcal{L}}{\partial \mathbf{y}_{2j-1}} 
\odot
\mathbf{y}_{2j-2}
.
\end{align}
The backpropagation through the circulant matrix consists of derivatives of  one-dimensional Fourier transform \cite{cheng2015exploration}, which is easier to write with explicit expression of FFT as shown in Eq. (\ref{eq:circ_matvec}):
\begin{align}
\begin{split}
\mathbf{F}_y &= \text{FFT}(\frac{\partial \mathcal{L}}{\partial \mathbf{y}_{2j}})
= \{\sum_{q=0}^{d-1} \frac{\partial \mathcal{L}}{\partial \mathbf{y}_{2j}}^{q} e^{-i2\pi \frac{p}{d} q}\}_{p=0}^{d-1}
,
\\
\frac{\partial \mathcal{L}}{\partial \mathbf{y}_{2j-1}}
&=
\text{IFFT}(\text{FFT}(\hat{\mathbf{c}}_{2j})\odot\mathbf{F}_y)
,
\\
\frac{\partial \mathcal{L}}{\partial \mathbf{c}_{2j}}
&=
\text{IFFT}(\text{FFT}(\hat{\mathbf{y}}_{2j-1})\odot\mathbf{F}_y)
.
\end{split}
\end{align}
Note that $\hat{\mathbf{c}}_{2j}$ and $\hat{\mathbf{y}}_{2j-1}$ are shifted from existing $\mathbf{c}_{2j}$ and $\mathbf{y}_{2j-1}$, respectively.
According to \cite{cheng2015exploration}, the shift pattern is fixed, i.e., from $(0,1,\dots,d-1)$ to $(0,d-1,\dots,1)$.

However, we find that this derivation is not efficient due to the need of applying FFT on shifted vectors. 
When the input is a real vector for FFT, it can be easily proved that the FFT on the shifted vector is equal to the conjugate of the result of FFT on original vector.
Therefore, we further improve the backward step as following:
\begin{align}
\begin{split}
\frac{\partial \mathcal{L}}{\partial \mathbf{y}_{2j-1}}
&=
\text{IFFT}(\text{conj}(\mathbf{F}_{2j})\odot\mathbf{F}_y)
,
\\
\frac{\partial \mathcal{L}}{\partial \mathbf{c}_{2j}}
&=
\text{IFFT}(\text{conj}(\mathbf{F}_{2j-1})\odot\mathbf{F}_y)
,
\end{split}
\end{align}
where the $\text{conj}(\cdot)$ means taking the conjugate of input vector. 
In this way, we show that the backward step can reuse some  results from the forward step.
Since FFT operation is the major computation complexity, it can be noticed that forward step takes 2 FFT and 1 IFFT, while backward step takes 2 IFFT and 1 FFT. 
Overall, both forward and backward steps have similar computation complexity.

\subsection{Block Partition}
The product of interleaved circulant and diagonal matrices inherently forms only square matrices. 
As diagonal matrix is always a square matrix,
to overcome the limitation, we decide to adopt the block-wise partitioning strategy for circulant matrices \cite{ding2017circnn}. 
This approach ensures compatibility with non-square weight matrices while preserving computational efficiency and storage benefits, maintaining the accuracy requirements of large-scale models.

In essence, this method partitions a non-square matrix into multiple square submatrices of equal size. If a dimension is not evenly divisible by the block size, it is automatically padded through replication. Each resulting square submatrix corresponds to a circulant matrix.
Formally, given a matrix $\mathbf{C}\in\mathbb{R}^{n\times n}$, we partition it into blocks of size $p$, resulting in $q_1=\lceil d_1/p \rceil$ blocks along the 
$d_1$-dimension and $q_2=\lceil d_2/p \rceil$ blocks along the 
$d_2$-dimension. Consequently, the original matrix is decomposed into 
$q_1 \times q_2$ smaller circulant matrices. During matrix multiplication, block-wise multiplication is applied, where corresponding submatrices are multiplied element-wise. This effectively transforms the non-square matrix multiplication into multiple independent square circulant matrix multiplications, allowing us to fully leverage the computational properties of circulant matrices. The block-wise matrix-vector multiplication can then be formulated as:
\begin{align}
\label{eq:blockcirc_matvec}
\begin{split}
\mathbf{h} 
&=
\mathbf{C}\mathbf{x}
= \{\mathbf{h}_i\}_{i=0}^{q_1-1}
,
\\
\mathbf{h}_i
&=
\sum_{j=0}^{q_2-1}
\mathbf{C}_{i,j}\mathbf{x}_j
\\
&=
\sum_{j=0}^{q-1}
\text{IFFT}(\text{FFT}(\mathbf{c}_{i,j}) \odot \text{FFT}(\mathbf{x}_j))
\\
&=
\text{IFFT}(\sum_{j=0}^{q-1}
\text{FFT}(\mathbf{c}_{i,j}) \odot \text{FFT}(\mathbf{x}_j)),
\end{split}
\end{align}
Each block matrix $\mathbf{C}_{i,j}$ represents a submatrix of the partitioned matrix, where $i$ and $j$ are integers ranging from $0$ to $(q_1-1)$ and $0$ to $(q_2-1)$, respectively. The vector $\mathbf{c}_{i,j}\in\mathbb{R}^{p\times 1}$ corresponds to each small circulant matrix.

It is worth mentioning that this method was specifically proposed to improve the generalization of our method. 
We only need to enable the partition for the first circulant matrix such that the rest matrices can always be square for maximum efficiency. 
However, even in scenarios where weight matrices are already square, this method also has the potential to further enhance model performance since there are more learnable parameters after partitioning. 
In our experiments with RoBERTa and ViT, all fine-tuned weight matrices were inherently square. Therefore, we set $p=d=768$ to achieve maximum efficiency. For experiments with LLaMA model, there are non-square weight matrices that we need to set different partition sizes to fit the training.

\subsection{Complexity Analysis}
\begin{figure}[!t]
\centering
\subfloat[ Adapter complexity on RoBERTa-base model.]{ 
\includegraphics[width=0.48\textwidth]{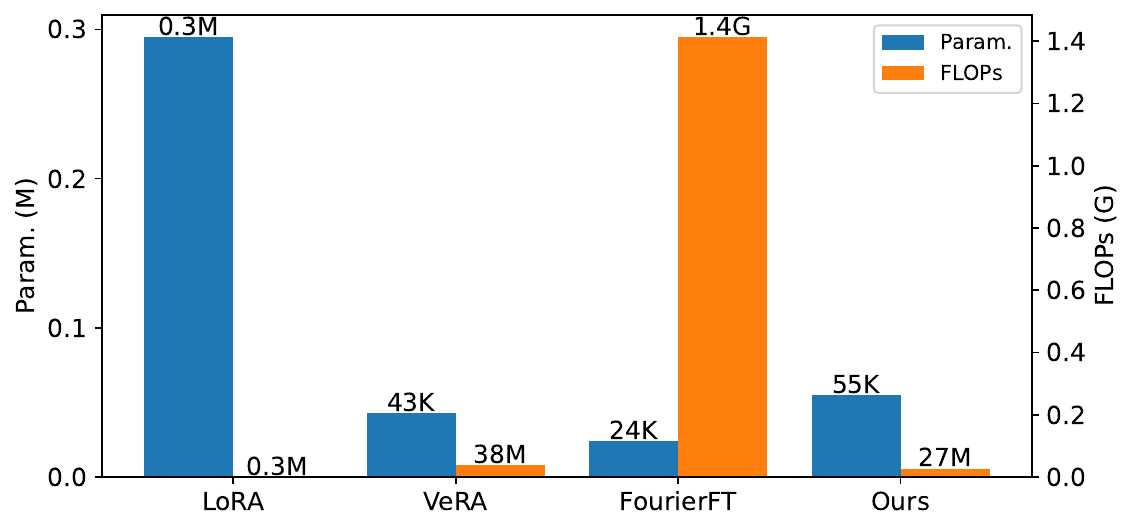}
}
\newline
\subfloat[ Adapter complexity on ViT-base model.]{
\includegraphics[width=0.48\textwidth]{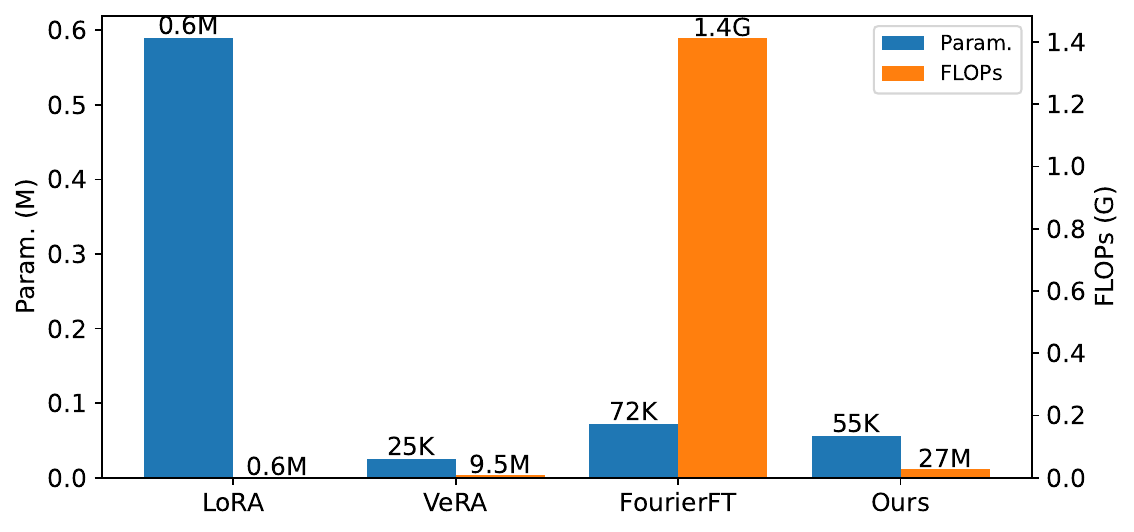}
}
\caption{
Complexity Analysis of different adapters.
FF method is not presented due to its high cost in both parameters and FLOPs. 
Our proposed block circulant adapters can balance between parameter amount and FLOPs. 
}
\label{fig:param_flops}
\end{figure}
Assume that the number of layers to be fine-tuned is \(L_t\) and \(\boldsymbol{\Delta} \mathbf{W}\in\mathbb{R}^{d\times d}\)(Priority is primarily given to cases where the weight matrix is square.)

\textbf{Parameters.} The number of parameters $\Theta$ to be trained for LoRA is given by \(|\Theta|_{\text{LoRA}} = 2 \times d \times L_t \times r\), where $|\cdot|$ means the cardinality. 
For VeRA, the total number of trainable parameters is \(|\Theta|_{\text{VeRA}} = (r+d) \times L_t\). However, in practice VeRA method can take large $r$ to achieve good performance.
For FourierFT, let number of spectral coefficients be \(n\), and the total number of trainable parameters is \(|\Theta|_{\text{FourierFT}} = n \times L_t\). 
For CDVFT(Ours), in the case of square weight matrices, the circulant matrix block size be \(p=d\). Assuming that the total number of circulant matrices and diagonal matrices is \(2m-1\), the total number of trainable parameters is \(|\Theta|_{\text{CDVFT}} =  (2m-1) \times d \times L_t\)
Fig. \ref{fig:param_flops} shows that Fourier domain based method, i.e., FourierFT and ours, require much less number of parameters than LoRA. 

\textbf{FLOPs.}
Fig.\ref{fig:param_flops} also analyzes the computational complexity.
It is important to note that computation complexity of FourierFT is independent of its parameter amount \(n\) since it always use 2D FFT to reconstruct \(\boldsymbol{\Delta} \mathbf{W}\). 
The complexity of our CDVFT is related to total number of circulant matrices and diagonal matrices and the circulant matrix block size, i.e., \(2m-1\) and $p$ .
The computational complexity of CDVFT is smaller than FourierFT. 
The main reason for the complexity difference is that in FourierFT, the computational complexity of the 2D FFT for computing \(\boldsymbol{\Delta} \mathbf{W}\) is \(O(d^2 log(d^2)))\). 
In CDVFT, the complexity brought by element-wise product and 1D FFT is \(O(mdlog(d))\), which significantly reduces the computational complexity while keeping similar number of training parameters. 
It can be seen that across all models, LoRA has the largest parameter amount, and FourierFT has the largest FLOPs.  
The complexity of the different models listed in Table \ref{tbl:llama_mt_gsm8k} verifies the analysis. 

\textbf{The case of non-square matrices.}
When the weight matrices in the fine-tuning layer are non-square, a block-wise partitioning strategy is required to ensure proper fine-tuning. In this case, the number of parameters and FLOPs depend on the block size $p$ of the circulant matrix.
A smaller $p$ results in more circulant matrix blocks, increasing the number of parameters required for representation and making the computation more complex. Therefore, in our experiments, we prioritize choosing a larger $p$. To validate our reasoning, we present results under different values of $p$ in Table \ref{tbl:llama_mt_gsm8k}. Furthermore, our approach achieves significantly lower FLOPs than FourierFT and requires fewer trainable parameters than LoRA, striking an effective balance between computational efficiency and model compactness.

\begin{table*}[!tb]
\centering
\setlength{\tabcolsep}{8.5pt}
\resizebox{0.97\textwidth}{!}{
\begin{tabular}{l|c c c c c c c c}
\hline
Method &  SST-2 & MRPC & CoLA & QNLI & RTE & STS-B & Avg. \\
\hline
FF & $94.8$ & $90.2$ & $63.6$ & $92.8$ & $78.7$ & $91.2$ & 85.2 \\
LoRA  & $95.1_{\pm0.2}$ & $89.7_{\pm0.7}$ & $63.4_{\pm1.2}$ & $93.3_{\pm0.3}$ & $78.4_{\pm0.8}$ & $91.5_{\pm0.2}$ & 85.2 \\
VeRA &$ 94.6_{\pm0.1} $ &$ 89.5_{\pm0.5}$ &$ 65.8_{\pm0.8}$ &$ 91.8_{\pm0.2}$ &$ 78.7_{\pm0.7}$ &$ 90.7_{\pm0.2}$ & 85.2 \\
FourierFT &$ 94.2_{\pm0.3} $ &$ 90.0_{\pm0.8}$ &$ 63.8_{\pm1.6}$ &$ 92.2_{\pm0.1}$ &$ 79.1_{\pm0.5}$ &$ 90.8_{\pm0.2}$ & 85.0 \\                
$\text{Ours}_{p=768}$  &$94.4_{\pm0.5} $ &$ 90.2_{\pm0.3}$ &$ 64.5_{\pm1.2}$ &$ 92.2_{\pm0.2}$ &$ 78.7_{\pm1.2}$ &$ 90.5_{\pm0.2}$ & 85.1\\
\hline
\end{tabular}
}
\caption{The performance of LoRA, FourierFT and our CDVFT methods is reported by fine-tuning the RoBERTa base model on 6 datasets of the GLUE benchmark. 
The experiments report Matthew correlation coefficient (MCC) for CoLA, Pearson correlation coefficient (PCC) for STS-B, and accuracy (Acc.) for all remaining tasks. 
Following \protect\cite{FourierFT}, we also report the median result out of 5 runs, each with a different random seed. 
The best result for each dataset is highlighted in bold. 
Higher metric value means better model performance for all datasets. }
\label{tbl:rob_glue}
\end{table*}

\section{Experiments}
In this section, we evaluate our CDVFT method across different domains, i.e., natural language understanding (NLU) and computer vision (CV): (1) fine-tune the RoBERTa model \cite{RoBERTa} on the General Language Understanding Evaluation (GLUE) dataset \cite{glue}; (2) fine-tune the vision transformer model \cite{vit} for various image classification tasks across different domains; (3) fine-tuning the LLaMA2-7b model on the Alpaca and GSM8K datasets.

Our proposed CDVFT is also compared with different adapters: 
Traditional full fine-tuning (FF) updates all model parameters, achieving high accuracy but incurring significant computational costs.
Low-Rank Adaptation (LoRA) \cite{LoRA} is a widely adopted fine-tuning technique for large models. It decomposes the weight matrices into low-rank components, significantly reducing the number of trainable parameters while maintaining performance.
Vector-based Random Matrix Adaptation (VeRA) \cite{vera} extends LoRA by sharing the low-rank matrices across all layers and inserting two additional vectors after the decomposed matrices, further improving efficiency.
LaMDA \cite{lamda} is another LoRA-based method that freezes the first projection matrix (PMA) during fine-tuning, gradually freezes the second projection matrix (PMB) in the early training stages, and introduces a low-rank square matrix between them.
LaMDA++ \cite{lamda} further extends this approach by assigning different ranks to different fine-tuning layers.
FourierFT \cite{DBLP:conf/icml/GaoWCLWC024} is a state-of-the-art fine-tuning method that transforms frequency-domain data into trainable weight matrices via the Fourier transform. By training only in the frequency domain, it effectively reduces the number of trainable parameters. Notably, our approach also leverages the Fourier domain, but with lower FLOPs. This is achieved by employing a more efficient FFT operation in our equation (\ref{eq:circ_matvec}) rather than the 2D FFT computation used in FourierFT.

\subsection{Natural Language Understanding}
\textbf{Models and Datasets.}
We evaluate CDVFT on the GLUE benchmark dataset, which consists of a diverse range of NLP tasks, each representing a specific type of language understanding task. 
These tasks include question answering, sentiment analysis, textual entailment, etc.  
Following the experiment setting as in \cite{FourierFT}, fine-tuning process runs on following tasks: CoLA, Corpus of Linguistic Acceptability \cite{CoLA}, which determines whether sentences adhere to grammatical rules; SST-2, Stanford Sentiment Treebank \cite{SST-2}, which classifies the sentiment of sentences as positive or negative; MRPC, Microsoft Research Paraphrase Corpus \cite{MRPC}, which assesses whether two sentences convey the same meaning; STS-B, Semantic Textual Similarity Benchmark \cite{STS-B}, which measures the semantic similarity score between sentence pairs; QNLI, Question Natural Language Inference \cite{QNLI}, which evaluates whether the second sentence correctly answers the question posed by the first; and RTE, Recognizing Textual Entailment \cite{RTE}, which identifies whether there is an entailment relationship between sentence pairs, functioning as a binary classification task.
RoBERTa base model \cite{RoBERTa} is a transformer based foundation model, which is widely used in natural language processing. 
It improves over existing under-trained BERT model \cite{BERT} while preserving the powerful attention mechanism. 
Thus, it is selected to serve as the foundation model for GLUE dataset. 

\textbf{Implementation Details.} 
Our CDVFT uses a total of 3 factor matrices, i.e., \(m = 2\). The block size of the circulant matrix is the same as the matrix size, i.e., $p=d=768$, and no block is performed.
It should be noted that only query and value weights in each transformer block are finetuned, which is also applied to LoRA, VeRA and FourierFT as in \cite{FourierFT}.
All implementations are in PyTorch \cite{paszke2019pytorch}.
It can be seen that the optimizer is AdamW \cite{adamw}.
For each dataset, there are different learning rates for foundation model language heads, query and value weight matrices. 
The scaling value is the $\alpha$ as in Eq. (\ref{eq.5}). 
The batch size and maximum input sequence length is set the same for all datasets. 

\textbf{Results.} Table \ref{tbl:rob_glue} summarizes fine-tuning results of all methods. 
The median metric value with standard deviation is reported out of 5 runs of experiments for each fine-tuning method, where each run takes a different random seed. 
The best performance for each dataset is highlighted in bold. 
Overall, compared with LoRA and FourierFT, our CDVFT method achieves  comparable or even better performance. 
Besides, according to Fig. \ref{fig:param_flops}, our CDVFT results in 5.33$\times$ less number of trainable parameters than LoRA and 51.89$\times$ less FLOPs than FourierFT while fine-tuning RoBERTa base model on GLUE dataset.

\begin{table}[tb]
\centering
\setlength{\tabcolsep}{3pt} 
\resizebox{0.48\textwidth}{!}{
\begin{tabular}{l | c c c c c}
\hline
Method & FLOPs & Param. & RESISC45 &  CIFAR100  & Avg.\\
\hline
FF & - & - & $96.1$ & $92.4$ & 94.2 \\
LoRA & $0.61M$ & $0.59M$ & $92.7$ & $92.0$ & $92.4$ \\
VeRA & $9.48M$ & $0.02M$ & $77.0$ & $84.8$ & $80.9$ \\
FourierFT & $1.41G$ & $0.07M$ & $92.0$ & $91.2$ & $91.6$\\
$\text{Ours}_{p=768}$ & $2.72M$ & $0.06M$ & $92.0$ & $91.1$ & $91.6$ \\
\hline
\end{tabular}
}
\caption{Fine-tuning results of the ViT Base model on different image classification datasets. The experiments report the accuracy (\%) after 10 epochs. 
}
\label{tbl:vit}
\end{table}

\subsection{Image Classification}
\textbf{Models and Datasets.} 
The experiment evaluates the performance of our CDVFT method in image classification tasks, utilizing the Vision Transformer (ViT) by \cite{vit} as the foundation model.
Following the setting in \cite{FourierFT}, we fine-tune on several challenging image classification datasets, only two are listed here for observation. 
RESISC45 \cite{resisc45} provides a diverse range of remote sensing images; 
CIFAR-100 \cite{cifar} is classical datasets of tiny images in  100 categories.

\textbf{Implementation details.}
We set \(m=2\) and $p=d=768$ for fine-tuning ViT base model across all these datasets. The ranks of LoRA and VeRA are 16 and 256 respectively. The n of FourierFT is set to 3000.
For all method, fine-tuning only runs the query and value weight matrices of ViT, which is the same as in \cite{FourierFT}.
The learning rate is set differently for fine-tuning ViT heads and query and value weight matrices.

\textbf{Results.} Table \ref{tbl:vit} presents the performance results on two image classification datasets after fine-tuning the ViT base model. Our CDVFT method demonstrates significant efficiency, requiring 10.7$\times$ fewer parameters than LoRA and 51.9$\times$ fewer FLOPs than FourierFT, while achieving similar or even better classification accuracy. Additionally, when compared to the latest fine-tuning method, VeRA, our approach uses fewer parameters. Although our method incurs a higher number of FLOPs than VeRA, it is important to note that VeRA's accuracy is relatively lower. To achieve comparable results to our method, VeRA would need to increase its rank, which would result in a corresponding increase in FLOPs.

\begin{table}[tb]
\centering
\setlength{\tabcolsep}{3pt} 
\resizebox{0.48\textwidth}{!}{
\begin{tabular}{c | c c c|c c c}
\hline
\multirow{2}{*}{Method} & \multicolumn{3}{c|}{MT-Bench}  & \multicolumn{3}{c}{GSM8K} \\
~ & FLOPs & Param. & Score &  FLOPs & Param. & Acc.\\
\hline
LoRA & $0.03G$ & $33.55M$ & $5.20$ & $0.01G$ & $28.05M$ & $36.9$ \\
VeRA & $2.29G$ & $1.65M$ & $5.08$ & - & - & - \\
FourierFT & $133.14G$ & $0.06M$ & $5.18$ & - & - & -\\
LaMDA & - & - & - & $0.06G$ & $4.37M$ & $37.9$ \\
LaMDA++ & - & - & - & - & $5.12M$ & $38.2$ \\                
$\text{Ours}_{p=2048}$ & $0.06G$ & $1.05M$ & $5.42$  & $0.22G$ & $7.26M$ & $37.8$ \\
$\text{Ours}_{p=4096}$  & $0.05G$ & $0.79M$ & $5.27$ & $0.17G$ & $4.51M$ & $37.9$ \\
\hline
\end{tabular}
}
\caption{Instruction tuning performance of LLaMA2-7B model. Higher accuracy and score value means better tuning  performance. Unavailable results are represented with ``-". For example, LaMDA++ is lack of rank information resulting in unknown FLOPs.
}
\label{tbl:llama_mt_gsm8k}
\end{table}

\subsection{Instruction Tuning}
\textbf{Models and datasets.} 
Instruction tuning is a training method that enables models to learn how to perform tasks based on natural language instructions. Its primary goal is to enhance the versatility of models in handling a wide range of tasks, allowing for improved understanding and execution of natural language commands. Compared to traditional supervised learning, instruction tuning emphasizes the model's broad adaptability to task requirements and can be applied in scenarios such as question-answering systems, multi-task learning, and natural language generation.

In this experiment, FourierFT and CDVFT are used to fine-tune LLaMA2-7B on the Alpaca dataset. Specifically, LLaMA2-7B is part of the LLaMA (Large Language Model Meta AI) series developed by Meta, featuring significant improvements in performance and scalability over the first generation. This model generates high-quality text while utilizing fewer computing resources. The Alpaca dataset is developed based on the Stanford Alpaca project and expanded from instruction data generated by OpenAI's GPT model, focusing on tasks related to natural language understanding and generation. 
The experiment generates answers to predefined questions from MT-Bench, evaluated using GPT-4. MT-Bench is a benchmark tool specifically designed to assess multi-task language models, quantifying their generalization ability, speed, and accuracy by comparing performance across multiple tasks.

Additionally, the GSM8K dataset is used for task-specific fine-tuning and evaluation of the model. GSM8K consists of approximately 8,000 math problem-solving instances, aimed at improving the model's ability to perform complex reasoning tasks. The dataset includes problem descriptions and detailed solutions, enabling the model to learn how to solve complex mathematical problems.

\textbf{Implementation details.} 
For FourierFT, the experiments follow previous work and adopt the configuration $n = 1000$.
For LoRA, $r = 8$ on the Alpaca dataset and $r = 64$ on the GSM8K dataset.
For VeRA and LaMDA, $r$ is $1024$ and $32$ respectively.
For CDVFT, the experiments still set $m = 2$, and the partition size $p$ is set as large as possible to achieve the minimum storage and computational cost.
Therefore, $p$ is set to $4096$ and $2048$.

Following \cite{DBLP:conf/icml/GaoWCLWC024}, we apply block circulant fine-tuning on query and value weight matrices inside the attention layer of two RoBERTa models and the LLaMA2-7B model fine-tuned on the alpaca dataset. (Except VeRA fine-tune on all layers in MHSA and MLP)
Following \cite{lamda}, we fine-tune on the MHSA and FFN layers of LLaMA2-7B model on the GSM8K dataset.
The classification head is fully fine-tuned. 

We evaluate the fine-tuned model on the Alpaca dataset using MT-Bench \cite{mtbench}, with GPT-4 \cite{gpt4} subsequently assigning scores to the model's responses for 80 multi-turn questions on a scale of 10.

\textbf{Results.}
Table \ref{tbl:llama_mt_gsm8k} shows the performance of the fine-tuned LLaMA2-7b model on the Alpaca and GSM8K datasets. 
It can be noticed that increasing partition size results in decreasing parameter amount and FLOPs.
However, the task score or accuracy does not always decrease with larger partition size. 
This may be caused by the over-parameterization of the large LLaMA model or  the strong structure of the proposed method that may serve as a regularization. 
Compared with other adapters, our method can balance between parameters amount and FLOPs.

\section{Conclusion}
Motivated by the recent success in Fourier domain based fine-tuning method, this paper proposes the CDVFT method that also learns parameters in Fourier domain. 
In particular, our method results in both trainable parameters savings and FLOPs reduction when compared with existing methods. 
The downstream task performance of our fine-tuned model achieves similar performance and sometime even better results across both natural language understanding and computer vision applications. 
These results effectively demonstrate the promising potential of our method and also the Fourier domain based fine-tuning methods. 
\newpage

\appendix

\section*{Acknowledgments}

This work was ﬁnancially supported by the National Key R\&D Program of China (Grant No. 2024YFA1211400).

\section*{Contribution Statement}
Siyu Liao and Zhongfeng Wang are co-corresponding authors. 

\bibliographystyle{named}
\bibliography{ijcai25}

\end{document}